\documentclass{article} 
\usepackage[preprint]{neurips_2023}

\usepackage[utf8]{inputenc} 
\usepackage[T1]{fontenc}    
\usepackage{hyperref}       
\usepackage{url}            
\usepackage{booktabs}       
\usepackage{amsfonts}       
\usepackage{nicefrac}       
\usepackage{microtype}      
\usepackage{xcolor}         
\usepackage{natbib}
\usepackage{subcaption}

\usepackage{tikz}
\usepackage{multirow}
\usepackage{amsmath}
\newcommand{\R}{\mathbb{R}}

\graphicspath{figures}

\title{Detecting and Mitigating Indirect\\ Stereotypes in Word Embeddings}

\author{Erin George \\
Department of Mathematics\\
University of California, Los Angeles\\
\texttt{egeo@math.ucla.edu}
\And Joyce Chew \\
Department of Mathematics\\
University of California, Los Angeles\\
\texttt{joycechew@math.ucla.edu} \And Deanna Needell  \\
Department of Mathematics\\
University of California, Los Angeles\\
\texttt{deanna@math.ucla.edu}
}

\newcommand{\sA}{A}
\newcommand{\sB}{B}
\newcommand{\sX}{X}
\newcommand{\sY}{Y}

\newtheorem{example}{Example}

\begin{document}

\maketitle

\begin{abstract}
Societal biases in the usage of words, including harmful stereotypes, are frequently learned by common word embedding methods.  These biases manifest not only between a word and an explicit marker of its stereotype, but also between words that share related stereotypes.  This latter phenomenon, sometimes called ``indirect bias,'' has resisted prior attempts at debiasing.  In this paper, we propose a novel method called Biased Indirect Relationship Modification (BIRM) to mitigate indirect bias in distributional word embeddings by modifying biased relationships between words before embeddings are learned.  This is done by considering how the co-occurrence probability of a given pair of words changes in the presence of words marking an attribute of bias, and using this to average out the effect of a bias attribute.  To evaluate this method, we perform a series of common tests and  demonstrate that measures of bias in the word embeddings are reduced in exchange for minor reduction in the semantic quality of the embeddings.  In addition, we conduct novel tests for measuring indirect stereotypes by extending the Word Embedding Association Test (WEAT) with new test sets for indirect binary gender stereotypes.   With these tests, we demonstrate the presence of more subtle stereotypes not addressed by previous work.  The proposed method is able to reduce the presence of some of these new stereotypes, serving as a crucial next step towards non-stereotyped word embeddings.
\end{abstract}

\section{Introduction}

Distributional word embeddings, such as Word2Vec~\citep{mikolov2013efficient} and GloVe~\citep{pennington2014glove}, are computer representations of words as vectors in semantic space.  These embeddings are popular because the geometry of the vectors corresponds to semantic and syntactic structure~\citep{mikolov2013linguistic}.  Unfortunately, societal stereotypes, such as those pertaining to race, gender, national origin, or sexuality, are typically reflected in word embeddings~\citep{bolukbasi2016man,caliskan2017semantics,garg2018word,papakyriakopoulos2020bias}. These stereotypes are so pervasive that they have proved resistant to many existing debiasing techniques~\citep{gonen2019lipstick}.
Techniques attempting to remove or mitigate bias in word vectors are common in the literature.  The typical case study for bias mitigation methods in the literature is binary gender.
Subspace methods, such as hard debiasing from~\citet{bolukbasi2016man} and GN-GloVe from~\citet{zhao2018learning}, attempt to identify or create a vector subspace of gender-related information (typically a ``gender direction'') and drop this subspace.  Counterfactual Data Substitution from~\citet{maudslay2019s}, based on~Counterfactual Data Augmentation from~\citet{lu2020gender}, swaps explicitly gendered words to counter stereotyped associations.  \citet{james2019probabilistic} and~\cite{qian2019reducing} both propose methods to reduce bias towards binary gender by encouraging learned conditional probabilities of words appearing with ``he'' and with ``she'' to be equal.

\citet{gonen2019lipstick} showed that common ``debiasing'' methods failed to meaningfully reduce bias in word embeddings.  They describe how bias can manifest not only as an undesirable association between stereotyped words and words marking the attributed associated with the streotype (``bias attribute'', for short), but also between stereotyped words themselves.  These manifestations are sometimes called \emph{direct bias} and \emph{indirect bias} (often also called ``explicit'' and ``implicit'' bias)
using the terminology introduced by~\cite{bolukbasi2016man}.  An example of this second manifestation of bias is that the word ``doctor'' might be associated more strongly with
stereotypically masculine 
words than with stereotypically feminine words.  At the time, bias mitigation algorithms commonly attempted to address direct bias while leaving indirect bias mostly present.

A common trend in the study of the indirect bias is the departure from \emph{stereotypes} as the object of study in favor of \emph{clustering}.  While the measures introduced by~\citet{bolukbasi2016man,caliskan2017semantics} attempt to quantify the existence of commonly understood stereotypes, work on indirect bias typically uses the measures introduced by~\citet{gonen2019lipstick} which merely attempt to measure how well proposed bias mitigation methods disperse words with similar relationships to the bias attribute in the embedding space.  These clustering measures, while useful at capturing some forms of indirect bias, are limited.  In particular, it is unclear how dispersed stereotyped words \emph{should} be in the embedding space, given that the stereotype of a word is not entirely arbitrary and can potentially be estimated based on its semantic, non-stereotypical, meaning.

These new bias measures have inspired countless new bias mitigation methods.  Nearest neighbor bias mitigation from~\citet{james2019probabilistic} attempts to equalize each word's association with its masculine (defined by the original undebiased embeddings) neighbors and its feminine neighbors.  Double hard debias from~\citet{wang2020double} projects off the direction defined by the ``most gender-biased words'' (again, based on alignment in the original embedding's ``gender direction'') in addition to the standard gender-related subspace.  \citet{bordia2019identifying} modify the loss function when learning word embeddings to penalize neutral words having large components in the gender-related subspace which can then be dropped off.  \citet{kumar2020nurse} propose RAN-Debias which attempts to disperse words in the embedding space that share similar binary gender biases (defined again by the original word embeddings) while preserving the original geometry as much as possible.  \citet{lauscher2020general} describe multiple bias mitigation methods: the standard projection method, averaging original word vectors with an orthogonal transformation that attempts to swap the bias attribute, and a neural method that uses a loss function to group together words exhibiting a bias attribute away from neutral words.  These methods, similarly to the bias measures of \citet{gonen2019lipstick}, focus on the clustering and dispersion of words in relation to the bias attribute.

In current state-of-the-art models, word embeddings have largely been replaced by contextualized embeddings from transformer models such as BERT~\citep{devlin2018bert} and GPT~\citep{radford2018improving}.  However, word embeddings remain a popular object of study when quantifying bias in NLP, in part due to their simplicity and theoretical results that make them easier to reason about.   As advances in the understanding of bias and stereotypes in word embeddings have been adapted for these newer models~\citep{liang2020towards,may2019measuring}, novel techniques to measure and mitigate bias in word embeddings remain relevant.

In this work, we use the phrase ``binary gender'' to refer to the common yet unrealistic simplification of gender as just ``male'' or ``female'', which we take to be the main bias attribute of study in this work.  This is a limitation of our work we hope to extend beyond in the future.  In addition, following the recommendations of~\citet{devinney2022theories}, we use the terms ``masculine'' and ``feminine'' instead of ``male'' and ``female'' in this work.

\section{Background}

\subsection{Word embeddings}

Many word embedding algorithms use the empirical probability that two given words appear near each other in the corpus~\citep{levy2014neural,pennington2014glove}.  This empirical probability is computed by counting how many times one word appears in the context of another as a \emph{word--context} pair.  A word--context pair is defined as a pair of words from the corpus that appear within a certain fixed distance from each other, the \emph{window size}, and within the same sentence.  A word--context pair designates one word as appearing in the context of another; in this paper, we will refer to a word--context pair as $(a,b)$ where $a$ is a the word appearing in the context of the word~$b$.  Contexts can be unidirectional or bidirectional.  In the unidirectional case, in a word--context pair~$(a,b)$, $a$ always occurs before $b$ in the corpus (or alternatively, always after).  In the bidirectional case, for any pair of nearby words $a$ and $b$, there are two word--contexts pairs: $(a,b)$ and $(b,a)$.

From these corpus statistics, word embedding algorithms learn vectors for each word in a way that enables the word co-occurrence statistics to be derived from the geometry of the vectors.  The exact details for how this is done is dependent on the exact word embedding algorithm used and is not important for this work.

\subsection{Word Embedding Association Test}

The Word Embedding Association Test (WEAT), introduced by~\citet{caliskan2017semantics}, is a common test used to quantify the presence of specific stereotypes in word embeddings.  Given two sets of target words $\sX$ and $\sY$ of equal size and two sets of attribute words $\sA$ and $\sB$, WEAT measures the association between the targets and the attributes.  The sets are chosen so that $\sX$ and $\sA$ are stereotypically linked with each other, and similarly for $\sY$ and $\sB$.

The association of a word $w\in \sX\cup\sY$ with the attributes is
\[s(w,\sA,\sB) = \frac{1}{|\sA|}\sum_{a \in \sA} \cos(\vec{w},\vec{a}) - \frac{1}{|\sB|}\sum_{b \in \sB} \cos(\vec{w},\vec{b})\]
where $\vec{u}$ denotes the word vector corresponding to the word $u$ and $\cos(\vec{u},\vec{v}) = \frac{\vec{u}\cdot\vec{v}}{\|\vec{u}\|\|\vec{v}\|}$
is the cosine similarity between $\vec{u}$ and $\vec{v}$.

The outputs of WEAT are the test statistic, $p$-value (of a permutation test), and effect size defined by the following equations respectively:
\begin{align*}
s(\sX,\sY,\sA,\sB) = \sum_{x \in \sX} s(x,\sA,\sB) &- \sum_{y \in \sY} s(y,\sA,\sB), \quad p = P(s(\sX_i,\sY_i,\sA,\sB) > s(\sX,\sY,\sA,\sB)), \\
\textrm{EffectSize}(\sX,\sY,\sA,\sB) &= \frac{\frac{1}{|\sX|}\sum_{x \in \sX} s(x,\sA,\sB) - \frac{1}{|\sY|}\sum_{y \in \sY} s(y,\sA,\sB)}{\operatornamewithlimits{stddev}_{w\in\sX\cup\sY}s(w,\sA,\sB)},
\end{align*}
where $(\sX_i, \sY_i)$ is a random partition of $\sX\cup\sY$ into two sets of equal size and $P$ denotes probability. 

Of these three measures, the effect size is most commonly used in the literature.  In this work, we report the effect size along with the $p$-value.  The effect size retains its original meaning from~\citet{caliskan2017semantics} of measuring the strength and direction of the tested stereotype.  We interpret the $p$-value not as a measure of statistical significance, as originally conceived.  Instead, we interpret it as another measure of the tested stereotype, indicating how easily words in $\sX$ can be separated from words in $\sY$ according to their association with $\sA$ and $\sB$: $p$-values of 0, 0.5, and 1 correspond to perfect separation according to the stereotype, no separation, and perfect separation according to the opposite stereotype, respectively.  We will interpret these as relative measures: a WEAT effect size closer to zero and a WEAT $p$-value closer to 0.5 both suggest reduced presence of the stereotype in the embeddings.

\section{Indirect Stereotypes}
\label{sec:indirect_stereo}

Previous work on measuring stereotypes has typically focused on measuring associations between stereotyped words and explicit markers of the stereotype, such as names
 and semantically gendered words.  In turn, previous work on mitigating bias has used these same markers as input for bias mitigation methods. By looking at the same sets of words for quantifying and mitigating bias, it is easy to overestimate the effect of bias mitigation.  This is what led \citet{gonen2019lipstick} to propose their new bias measures.  These are based around measuring how clustered previously biased words are, but it is unclear what result should be desired from these measures.  To address this issue, we demonstrate that we can capture some forms of the remaining bias as stereotypes, which we refer to as \emph{indirect stereotypes}.

\begin{figure}[h]
    \centering
    \begin{tikzpicture}
\draw[black, ultra thick, <->] (-0.1788854,3.910557) -- (-3.821115,2.0895528);
\draw[black, ultra thick, <->] (0,3.8) --  (0,2.2);
\draw[black, ultra thick, <->] (0.1788854,0.0894428) --  (3.821115,1.910557);
\draw[gray, ultra thick, <->, dashed] (-3.6,2) --  (-0.2,2) node[midway, above]{intensified} node[midway,below]{relationship};
\draw[gray, thick, <->, dashed] (0.2,2) --  (3.6,2) node[midway, above]{suppressed} node[midway,below]{relationship};

\filldraw[black]  (0,4) circle (2pt) node[anchor=south]{man};
\filldraw[black]  (0,2) circle (2pt) node[anchor=north]{engineer};
\filldraw[black] (-4,2) circle (2pt) node[anchor=north east]{handsome};
\filldraw[black]  (4,2) circle (2pt) node[anchor=north west]{sentimental};
\filldraw[black]  (0,0) circle (2pt) node[anchor=north]{woman};
\end{tikzpicture}

    \caption{An example showing indirect stereotypes.
    }
    \label{fig:indirect_bias}
\end{figure}
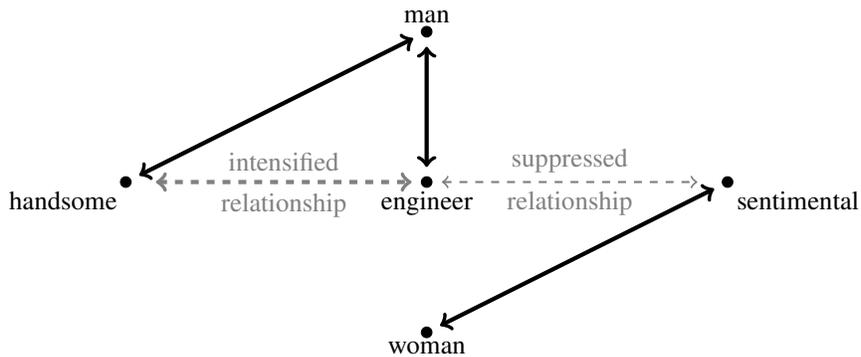

An example of an indirect stereotype is shown in Figure~\ref{fig:indirect_bias}.  There are three words here exhibiting a stereotype in regards to binary gender: ``handsome'' and ``engineer,'' which are typically both masculine and not feminine, and ``sentimental,'' which is typically feminine and not masculine.  In a corpus with binary gender stereotypes, we would expect a quantifiably stronger association between ``handsome'' and ``engineer'' than between ``sentimental'' and ``engineer'' to exist purely because they are both stereotypically used to refer to men,
even if there is not a stereotype that engineers are more handsome than they are sensitive.

These stereotypes can come about as a result of sentences that exhibit multiple stereotypes at the same time.  For example, the sentence ``he is an a handsome engineer'' exhibits the masculine stereotypes for ``handsome'' and ``engineer''.  In a corpus that exhibits both of these stereotypes, this sentence would be more likely to occur than the following related sentences: ``She is a handsome engineer," ``He is a sentimental engineer," ``She is a sentimental engineer." 
This will result in ``handsome'' occurring with ``engineer'' more frequently than ``sentimental'' does.  However, when word embeddings are ``debiased'' with respect to binary gender, the goal in previous work is typically to obtain word vectors that would predict each pair of the previous examples that differ only in choice of pronoun as equally likely to occur.  This is not enough to correct the associations between ``handsome,'' ``engineer,'' and ``sentimental.''  Furthermore, the association between ``engineer'' and ``handsome'' should exist even in sentences without explicit reference to binary gender, as in those cases ``engineer'' is more likely to refer to a man than a woman.  More details on this analysis can be found in Appendix~\ref{sec:indirect_analysis}.

To test indirect stereotypes, we use WEAT with the categories that are stereotyped according to binary gender (these are science/art, math/art, and career/family), as well as a test set of professions and adjectives that are stereotypically masculine and feminine.  These new test sets and the process used to generate them are detailed in Appendix~\ref{sec:word_lists}.  While we will use WEAT as our primary test for stereotypes in word embeddings, subsequent work has called attention to shortcomings of the measure.  We give a more thorough discussion with our analysis, at the end of Section~\ref{sec:results}.

\section{Bias Mitigation Method}
\label{sec:method}

In light of the issues discussed in Section~\ref{sec:indirect_stereo}, we propose a method to correct word co-occurrence probabilities, called Biased Indirect Relationship Modification (BIRM).  Instead of only looking at the co-occurrence probability of two words, we consider the probability a word co-occurs with another and one of the two also occurs near a word marking the bias attribute.  By using the word marking a bias attribute as a proxy for the bias attribute itself, we determine how the relationship between the two words varies as the bias attribute varies.  Then, we can approximate what the relationship between two words would be if one of them had no association with the bias attribute.

\begin{example}\label {ex:engineer_handsome}
Suppose we would like to mitigate one of the stereotypes in Figure~\ref{fig:indirect_bias}, perhaps that ``engineer'' and ``handsome'' are more closely associated than they should be based on non-stereotypical semantics.  In the corpus, it might be that an engineer is described as handsome three times as often when ``engineer'' refers to a man as opposed to when it refers to a woman.  It might also be the case that ``engineer'' refers to a man twice as often as it refers to a woman.  If ``engineer'' is adjusted so that it refers to men just as often as women, then the probability that an engineer is referred to as handsome should also be adjusted to be
\begin{equation}
    \frac{\tfrac{1}{2}\cdot P(\text{``handsome''}|\text{man}) + \tfrac{1}{2}\cdot P(\text{``handsome''}|\text{woman})}{\tfrac{2}{3}\cdot P(\text{``handsome''}|\text{man}) + \tfrac{1}{3}\cdot P(\text{``handsome''}|\text{woman})} = \frac{6}{7}
    \label{eq:engineer_handsome}
\end{equation}
of what it was before, at least when just considering occurrences where ``engineer'' refers to a man or woman.  Note that we use probabilistic notation here to refer to frequencies, as if a word--context pair is selected uniformly at random.

\end{example}

The general method is as follows.
Consider two words~$a$ and~$b$ where at least one of~$a$ or~$b$ is a word to be neutralized with respect to this bias.  For each word--context pair, we define a score variable $s$ that takes values in some finite set $S$.

The probability
a randomly selected word--context pair is~$(a,b)$ can  be decomposed as
 \begin{align*}
 P(a, b)
    &= \sum_{x \in S} P(a | b, s=x)P(s=x | b)P(b)
\end{align*}
using the definition of conditional probability and where $P(a | b)$ denotes the probability the word $a$ is the first word in the pair, conditioned on the context in the pair being word $b$.  This representation isolates the contribution of the bias attribute on $P(a, b)$.  The term $P(s=x | b)$ indicates the relation $b$ has with the bias attribute, while the term $P(a | b, s=x)$ describes how the relationship between $a$ and $b$ is influenced by the bias attribute.

Now if~$b$ is replaced with a hypothetical neutral word~$b'$ that has the same meaning except with no association to the bias attribute, the probability of~$b'$ occurring with~$a$ can be approximated.  First, we make note of the following approximate equalities:
\begin{align*}
    P(a | b', s=x) \approx P(a | b, s=x), \quad
    P(s = x | b') \approx P(s = x), \quad
    P(b') \approx P(b).
\end{align*}
In the first line, the explicit presence of the bias attribute of interest is likely to overpower the bias $b$ has.  In the second line, $b'$ has no bias and should there be independent of the bias attribute.  In the third line, $b$ and $b'$ have the same meaning and should therefore have similar co-occurrence probabilities. These relations imply
 \begin{equation}\label{eq:neutralize_b}
 P(a, b')
    \approx \sum_{x \in S} P(a | b, s=x)P(s=x)P(b).
\end{equation}
Reconsider Example~\ref{ex:engineer_handsome}.  Suppose our score $s$ perfectly marks whether ``engineer'' refers to a man or a woman, taking values of 0 for neither, -1 for man, or 1 for woman.  If the two gendered values are equally likely, i.e., $P(s=-1) = P(s=1)$, we see that \eqref{eq:neutralize_b} is precisely the numerator of~\eqref{eq:engineer_handsome}, multiplied by some constant and with an added term for $s=0$ to compensate for the instances where ``engineer'' refers to neither a man nor woman.  With the switch from semantics to co-occurrence, this is now an approximation.

The previous example suggests that we may want to set
\[P(s=1) = P(s=-1) =\frac{1}{2}(P(s=1) + P(s=-1))\]
to avoid the assumption that $s=1$ and $s=-1$ are equally likely in the corpus.  Indeed, this modification can easily be made to the method, and would likely help to correct an overall corpus bias of binary gender.  However, it is not clear how sensible this modification is for bias involving a minority group, as in this situation an unequal distribution of $s$ may not represent a bias.  For this reason, we prefer the generality of~\eqref{eq:neutralize_b}.

This approximation is the basis of our proposed bias mitigation method.  We propose the right hand side of \eqref{eq:neutralize_b} can replace the probability of a word--context pair being~$(a,b)$ whenever this quantity would be used, where $a$ or $b$ is a word that is wanted to be neutralized.  This can be done easily with word embedding methods that directly use the probability or counts of a specific word--context pair occurring, such as GloVe or PPMI factoring, but in principle could be adapted even to methods that only indirectly use these probabilities, such as word2vec.

It remains to determine a score $s$ for each word--context pair in the corpus.  A simple way of doing this is to define sets of words that explicitly capture the bias attribute, and defining $s$ based on which of these words appear near the word--context pair.  For example, for binary gender, we could simply count the number of times ``he'' or ``she'' appears within some window of the word--context pair, and then define $s = 0$ if these are equal, $s = -1$ if ``he'' occurs more, or $s = 1$ if ``she'' occurs more.  This creates a score that does capture relevant information about the bias attribute, and is a good first step.  However, a significant issue with this method is that most word--context pairs will occur near neither ``he'' nor ``she'', so $s = 0$ will be over-represented.  We can mitigate this by using larger sets of words, but would be still limited by the reality that most word--context pairs do not occur near words that can explicitly mark the bias attribute.

An improvement to this method is, for each word $w$ in the corpus, to assign a score $\tilde{s}(w)$ for that word.  Then use the average of the scores of all the words appearing near the word--context pair to inform the score of the pair.  For example, in the case of binary gender, we could assign positive values to feminine words, negative values to masculine words, and zero to the rest of the words.  Then for each word--context pair, we sum the scores of the nearby words and set $s \in \{-1,0,1\}$ depending on whether this is negative, zero, or positive.  That is, for a given context $b$, the score of the word--context pair is
\begin{equation}\label{eq:score_sum}
s = \operatorname{sgn}\left(\sum_{w \in \mathrm{context}(b)} \tilde{s}(w)\right)
\end{equation}
where the sum ranges over all words that $b$ serves as a context for.  The advantage of this method is that we can determine the scores of most words algorithmically.

One general method of doing so is to start with explicit sets of words that mark the bias attribute, as in the previous method, and use these words to define a score for the rest based on the co-occurrence probabilities.  We find that, in the case of bias with respect to binary classes, these scores work well when chosen according to the log of odds of the word appearing near one set to the other, relative to the full corpus odds.  That is, given a word $w$ and two sets $A$ and $B$, we define the score of $w$ to be
\begin{equation}\label{eq:logrelodds}
\tilde{s}(w) = c \log \left(\frac{\left(\sum_{a\in A}p(w, a)\right) / \left(\sum_{b\in B}p(w, b)\right)}{\left(\sum_{a\in A}p(a)\right) / \left(\sum_{b \in B}p(b)\right)}\right)
\end{equation}
for some constant $c \in\R$. As words in $A$ and $B$ are chosen to directly capture the bias attribute, we can set their score to be a large (positive or negative) fixed value instead of the result of \eqref{eq:logrelodds}.  Note that for practical reasons, these scores are then rounded to the nearest integer, and the constant $c$ can be chosen to dictate the range of integer values the score can take as well as the likelihood that a word--context pair will receive a score of 0.

\section{Results}

\label{sec:results}

We conduct experiments by training GloVe embeddings on the UMBC webbase corpus~\citep{han2013umbc} and then attempting to mitigate the presence of bias from binary gender. We first test on real data and display semantic tests demonstrating low semantic degradation for the method, followed by WEAT tests for direct biases and finally for indirect biases. We also create a semi-synthetic dataset to further showcase the effects of the mitigation.

We compare the proposed method from Section~\ref{sec:method} with GloVe, with and without Counterfactual Data Subsitution~\citep{maudslay2019s}.  These are labeled as ``BIRM'', ``Original'', and ``CDS'' respectively in the following tables. All results are obtained by training word vectors ten times.  We report the minimums, medians, and maximums of the respective measures, in this order.  In all tables, the best results for each row are shown in bold.  CDS was chosen as the bias mitigation method for comparison as it is a pre-processing method, similar to BIRM.

For all methods, we train GloVe with its default hyperparameters.  That is, we use an embedding dimension of 300 and train for 15 epochs, and to determine the co-occurrence matrix we use a window size of 15 and skip over words with less than 5 occurrences in the corpus.  We also preprocess our corpus by removing capitalization, skipping sentences with less than five words, and using Stanford CoreNLP~\citep{manning-EtAl:2014:P14-5} to separate out certain lexemes from words (e.g., n't and 's).

For the proposed bias mitigation method, we use the following sets of words:
\begin{align*}
    A &= \{\text{``he'', ``him'', ``his'', ``himself'', ``man'', ``men'', ``boy'', ``boys''}\}\\
    B &= \{\text{``she'', ``her'', ``hers'', ``herself'', ``woman'', ``women'', ``girl'', ``girls''}\}
\end{align*}
and use \eqref{eq:score_sum} with \eqref{eq:logrelodds} and $c = 1.3$ to determine the scores.  We set the score of words in $A$ to be $-100$ and the score of words in $B$ to be $100$.  The hyperparameters for BIRM were determined by results for a single set of vectors, after which ten new sets of vectors were generated for robustness.

We train our vectors using an Intel Core i9-9920X CPU.  Preprocessing the corpus and generating corpus counts with and without the correction for BIRM was done in less than 24 hours.  Generating the synthetic dataset with corpus counts was done in less than 4 hours.  Processing the corpus with CDS was done in less than 72 hours.  Each set of vectors was trained for 150 minutes, for a total of 225 hours of compute time.

\begin{table}
\centering
\caption{Semantic measures}\label{tab:semantic}
\begin{subtable}[c]{0.475\linewidth}
\caption{Word similarity tasks}
\begin{tabular}{lrrrr}
\toprule
Test & Original && BIRM & CDS\\
\midrule
\multirow{3}{*}{\shortstack[l]{MEN}} & $0.707$ && $0.699$ & $\mathbf{0.702}$ \\
& $0.715$ && $\mathbf{0.715}$ & $0.714$ \\
& $0.718$ && $\mathbf{0.722}$ & $\mathbf{0.722}$ \\[2pt]
\multirow{3}{*}{\shortstack[l]{WS353}} & $0.527$ && $0.524$ & $\mathbf{0.527}$ \\
& $0.537$ && $\mathbf{0.540}$ & $0.535$ \\
& $0.546$ && $\mathbf{0.553}$ & $0.547$ \\[2pt]
\multirow{3}{*}{\shortstack[l]{WS353R}} & $0.479$ && $0.465$ & $\mathbf{0.481}$ \\
& $0.486$ && $0.487$ & $\mathbf{0.491}$ \\
& $0.502$ && $\mathbf{0.502}$ & $\mathbf{0.502}$ \\[2pt]
\multirow{3}{*}{\shortstack[l]{WS353S}} & $0.643$ && $0.638$ & $\mathbf{0.640}$ \\
& $0.653$ && $\mathbf{0.660}$ & $0.650$ \\
& $0.664$ && $\mathbf{0.672}$ & $0.660$ \\[2pt]
\multirow{3}{*}{\shortstack[l]{SimLex999}} & $0.363$ && $\mathbf{0.361}$ & $0.360$ \\
& $0.371$ && $\mathbf{0.370}$ & $0.365$ \\
& $0.378$ && $\mathbf{0.375}$ & $0.373$ \\[2pt]
\multirow{3}{*}{\shortstack[l]{RW}} & $0.349$ && $0.341$ & $\mathbf{0.347}$ \\
& $0.356$ && $\mathbf{0.367}$ & $0.356$ \\
& $0.363$ && $\mathbf{0.371}$ & $0.368$ \\[2pt]
\multirow{3}{*}{\shortstack[l]{RG65}} & $0.786$ && $0.774$ & $\mathbf{0.779}$ \\
& $0.796$ && $0.784$ & $\mathbf{0.792}$ \\
& $0.809$ && $0.792$ & $\mathbf{0.808}$ \\[2pt]
\multirow{3}{*}{\shortstack[l]{MTurk}} & $0.607$ && $0.597$ & $\mathbf{0.601}$ \\
& $0.614$ && $\mathbf{0.611}$ & $0.609$ \\
& $0.621$ && $\mathbf{0.634}$ & $0.616$ \\
\bottomrule
\end{tabular}
\end{subtable}
\begin{subtable}[c]{0.516\linewidth}
\caption{Categorization tasks}
\begin{tabular}{lrrrr}
\toprule
Test & Original && BIRM & CDS\\
\midrule
\multirow{3}{*}{\shortstack[l]{AP}} & $0.642$ && $0.622$ & $\mathbf{0.627}$ \\
& $0.655$ && $0.639$ & $\mathbf{0.654}$ \\
& $0.679$ && $0.654$ & $\mathbf{0.669}$ \\[2pt]
\multirow{3}{*}{\shortstack[l]{BLESS}} & $0.780$ && $\mathbf{0.805}$ & $0.785$ \\
& $0.833$ && $\mathbf{0.828}$ & $0.823$ \\
& $0.870$ && $\mathbf{0.845}$ & $0.840$ \\[2pt]
\multirow{3}{*}{\shortstack[l]{Battig}} & $0.409$ && $\mathbf{0.409}$ & $0.400$ \\
& $0.423$ && $\mathbf{0.418}$ & $0.415$ \\
& $0.427$ && $\mathbf{0.433}$ & $0.423$ \\[2pt]
\multirow{3}{*}{\shortstack[l]{ESSLI$_{2\mathrm{c}}$}} & $0.578$ && $\mathbf{0.578}$ & $0.533$ \\
& $0.600$ && $\mathbf{0.600}$ & $\mathbf{0.600}$ \\
& $0.667$ && $\mathbf{0.622}$ & $0.600$ \\[2pt]
\multirow{3}{*}{\shortstack[l]{ESSLI$_{2\mathrm{b}}$}} & $0.725$ && $\mathbf{0.725}$ & $\mathbf{0.725}$ \\
& $0.750$ && $\mathbf{0.750}$ & $\mathbf{0.750}$ \\
& $0.850$ && $\mathbf{0.850}$ & $0.825$ \\[2pt]
\multirow{3}{*}{\shortstack[l]{ESSLI$_{1\mathrm{a}}$}} & $0.773$ && $\mathbf{0.795}$ & $0.750$ \\
& $0.807$ && $\mathbf{0.818}$ & $\mathbf{0.818}$ \\
& $0.818$ && $\mathbf{0.864}$ & $0.841$ \\
\bottomrule
\end{tabular}
\caption{Analogy tasks}
\begin{tabular}{lrrrr}
\toprule
Test & Original && BIRM & CDS\\
\midrule
\multirow{3}{*}{\shortstack[l]{Google}} & $0.686$ && $0.680$ & $\mathbf{0.684}$ \\
& $0.689$ && $0.688$ & $\mathbf{0.690}$ \\
& $0.693$ && $0.692$ & $\mathbf{0.693}$ \\[2pt]
\multirow{3}{*}{\shortstack[l]{MSR}} & $0.652$ && $\mathbf{0.649}$ & $0.642$ \\
& $0.659$ && $0.652$ & $\mathbf{0.654}$ \\
& $0.664$ && $0.657$ & $\mathbf{0.660}$ \\[2pt]
\multirow{3}{*}{\shortstack[l]{SemEval2012$_2$}} & $0.174$ && $\mathbf{0.167}$ & $0.162$ \\
& $0.180$ && $\mathbf{0.176}$ & $0.172$ \\
& $0.183$ && $\mathbf{0.185}$ & $0.176$ \\
\bottomrule
\end{tabular}
\end{subtable}
\end{table}

As a first test, we evaluate all three word embeddings on a series of common semantic tests~\citep{jastrzebski2017evaluate}.  The minimum, median, and maximum results of these tests before and after the two tested bias mitigation methods are shown in Table~\ref{tab:semantic}.  For all tests, a higher score indicates a better representation of the tested semantics in the word embeddings.  We only bold the better result of the two bias mitigation methods, as we would expect any bias mitigation method to reduce the semantic quality of vectors.  The tests are grouped into three categories: word similarity tasks, categorization tasks, and analogy tasks.  Overall, neither BIRM or CDS consistently outperform the other on these tasks.  Further, neither method appears to have a significant decrease in semantic quality compared with original GloVe.  The drops of these scores are largest for the categorization tasks, but in these cases we see that CDS often has the larger degradation in accuracy.  BIRM sees the largest drop in accuracy for the RG65 and AP tests, but even for these tests the drops are relatively minor.

\begin{table}
\caption{$p$-values and effect sizes of WEAT for direct stereotypes}
\label{tab:WEAT_p_direct}
\centering
\begin{tabular}{lrrrrrrr}
\toprule
&\multicolumn{3}{c}{$p$-value} && \multicolumn{3}{c}{Effect size}\\
Test & Original & BIRM & CDS && Original & BIRM & CDS \\
\midrule
\multirow{3}{*}{\shortstack[l]{
Math/Art\\Masc./Fem. Words}} & $0.0161$ & $0.144$ & $\mathbf{0.754}$ && $0.565$ & $\mathbf{-0.0947}$ & $-1.064$\\
 & $0.0409$ & $\mathbf{0.364}$ & $0.919$ && $0.901$ & $\mathbf{0.200}$ & $-0.734$ \\
 & $0.144$ & $\mathbf{0.567}$ & $0.982$ && $1.075$ & $0.574$ & $\mathbf{-0.378}$\\[2pt]
\multirow{3}{*}{\shortstack[l]{Science/Art\\Masc./Fem. Words}} & $0.00466$ & $0.0411$ & $\mathbf{0.572}$ && $0.736$ & $\mathbf{-0.401}$ & $-1.110$  \\
 & $0.0420$ & $\mathbf{0.328}$ & $0.880$ && $0.891$ & $\mathbf{0.239}$ & $-0.623$ \\
 & $0.0817$ & $\mathbf{0.777}$ & $0.987$ && $1.117$ & $0.899$ & $\mathbf{-0.0961}$\\[2pt]
\multirow{3}{*}{\shortstack[l]{Masc./Fem. Names\\Career/Home}} & $\mathbf{7.8\mathrm{e}{-5}}$ & $\mathbf{7.8\mathrm{e}{-5}}$ & $\mathbf{7.8\mathrm{e}{-5}}$ && $1.726$ & $1.607$ & $\mathbf{1.442}$ \\
 & $\mathbf{7.8\mathrm{e}{-5}}$ & $\mathbf{7.8\mathrm{e}{-5}}$ & $\mathbf{7.8\mathrm{e}{-5}}$ && $1.765$ & $1.705$ & $\mathbf{1.595}$ \\
 & $7.8\mathrm{e}{-5}$ & $7.8\mathrm{e}{-5}$ & $\mathbf{2.33\mathrm{e}{-4}}$ && $1.788$ & $1.767$ & $\mathbf{1.652}$ \\
\bottomrule
\end{tabular}
\end{table}

To test the standard direct stereotypes towards binary gender, we use WEAT with three binary gender stereotypes used in the original paper.  The $p$-values and effect sizes for these tests are in Table~\ref{tab:WEAT_p_direct}.  All three of these stereotypes are present in the original GloVe embedding, with the association between traditionally masculine and feminine names with career and home words being the strongest.  The presence of all three stereotypes are reduced as a result of both bias mitigation methods, although the career/home stereotype is not substantially reduced by either method.  We focus our attention on the two direct tests that both methods perform well at.  We note that WEAT measures exhibit significance variance over the randomness used in training the word vectors.  The ranges these measures take for the original GloVe and BIRM do overlap, but not substantially.  Typically, WEAT indicates that CDS and BIRM exhibit less direct bias than the original GloVe vectors do.  The ranges for CDS and original GloVe do not overlap, but this is because the WEAT scores for these tests indicate that CDS typically \emph{overcorrects} the direct bias, sometimes yielding scores that show a stereotype in the reverse direction is encoded as strongly as the original.

\begin{table}
\caption{$p$-values and effect sizes of WEAT for indirect stereotypes}
\label{tab:WEAT_p_indirect}
\centering
\begin{tabular}{lrrrrrrr}
\toprule
&\multicolumn{3}{c}{$p$-value} && \multicolumn{3}{c}{Effect size}\\
Test & Original & BIRM & CDS && Original & BIRM & CDS \\
\midrule
\multirow{3}{*}{\shortstack[l]{
Professions\\Adjectives}} & $0.00150$ & $\mathbf{0.00933}$ & $0.00280$ && $0.926$ & $\mathbf{0.661}$ & $0.927$ \\
 & $0.00374$ & $\mathbf{0.0388}$ & $0.0101$ && $1.140$ & $\mathbf{0.810}$ & $1.029$\\
 & $0.0197$ & $\mathbf{0.0780}$ & $0.0194$ && $1.257$ & $\mathbf{1.021}$ & $1.194$ \\[2pt]
\multirow{3}{*}{\shortstack[l]{
Math/Art\\Adjectives}} & $1.55\mathrm{e}{-4}$ & $\mathbf{0.00443}$ & $7.8\mathrm{e}{-5}$ && $1.073$ & $\mathbf{1.033}$ & $1.311$ \\
 & $0.00136$ & $\mathbf{0.0103}$ & $7.77\mathrm{e}{-4}$ && $1.306$ & $\mathbf{1.137}$ & $1.382$ \\
 & $0.0176$ & $\mathbf{0.0203}$ & $0.00303$ && $1.511$ & $\mathbf{1.267}$ & $1.531$ \\[2pt]
\multirow{3}{*}{\shortstack[l]{
Science/Art\\Adjectives}} & $0.00218$ & $\mathbf{0.00373}$ & $04.66\mathrm{e}{-4}$ && $1.134$ & $\mathbf{1.051}$ & $1.250$   \\
& $0.00400$ & $\mathbf{0.0108}$ & $0.00330$ && $1.289$ & $\mathbf{1.152}$ & $1.346$  \\
 & $0.0121$ & $\mathbf{0.0196}$ & $0.00559$ && $1.367$ & $\mathbf{1.296}$ & $1.483$\\
\bottomrule
\end{tabular}
\end{table}

Lastly, we investigate the presence of indirect stereotypes as described in Section~\ref{sec:indirect_stereo}.  We use a list of stereotypically masculine and feminine adjectives from~\cite{hosoda2000current} and look at its association with stereotypically masculine and feminine professions from~\cite{caliskan2017semantics}, as well as the same math/art, science/art, and career/home categories from the standard WEAT tests.  The full word lists for our new tests are shown in Appendix~\ref{sec:word_lists}. The results of these experiments are in Table~\ref{tab:WEAT_p_indirect}.  From this, it can be seen that all these stereotypes do exist in the original word embeddings.  Counterfactual Data Substitution does not succeed at significantly reducing any of these stereotypes.  BIRM reduces the presence of all these stereotypes to a greater extent than CDS, although it is not uniformly successful.  In particular, it has modest success in reducing the presence of the indirect stereotypes with professions and has less success for the other two stereotypes.  There is significant variance in the WEAT measures and significant overlap in the range of scores for the other two stereotypes with all three methods.  However, we note that BIRM's performance is higher than the corresponding scores for original GloVe and CDS for all reported values.

\begin{table}
\caption{$p$-values and effect sizes of WEAT for synthetic direct and indirect stereotypes}
\label{tab:WEAT_synthetic}
\centering
\begin{tabular}{lrrrrr}
\toprule
&\multicolumn{2}{c}{$p$-value} && \multicolumn{2}{c}{Effect size}\\
Test & Original & BIRM && Original & BIRM \\
\midrule
\multirow{3}{*}{\shortstack[l]{
Synthetic Adjectives\\Masc./Fem. Words}} & $5\mathrm{e}{-6}$ & $\mathbf{0.000482}$  && $1.994$ & $\mathbf{0.243}$\\
& $5\mathrm{e}{-6}$ & $\mathbf{0.0218}$ && $1.997$ & $\mathbf{0.913}$  \\
 & $5\mathrm{e}{-6}$ & $\mathbf{0.308}$ && $1.999$ & $\mathbf{1.402}$ \\[2pt]
\multirow{3}{*}{\shortstack[l]{Synthetic Nouns\\Masc./Fem. Words}} & $5\mathrm{e}{-6}$& $\mathbf{0.00129}$ && $1.995$ & $\mathbf{0.192}$  \\
 & $5\mathrm{e}{-6}$& $\mathbf{0.0332}$  && $1.997$ & $\mathbf{0.828}$ \\
 & $5\mathrm{e}{-6}$& $\mathbf{0.349}$  && $1.999$ & $\mathbf{1.267}$ \\[2pt]
\multirow{3}{*}{\shortstack[l]{Synthetic Adjectives\\Synthetic Nouns}} & $\mathbf{5\mathrm{e}{-6}}$ & $\mathbf{5\mathrm{e}{-6}}$ && $1.998$ & $\mathbf{0.965}$ \\
 & $5\mathrm{e}{-6}$ & $\mathbf{0.000227}$ && $1.998$ & $\mathbf{1.410}$\\
 & $5\mathrm{e}{-6}$ & $\mathbf{0.0164}$ && $1.999$ & $\mathbf{1.817}$ \\
\bottomrule
\end{tabular}
\end{table}

To assess the performance of BIRM in a setting with pronounced indirect stereotypes, we augment 3 million sentences from the UMBC webbase corpus with 32,000 synthetic sentences. Each sentence is of the form “[pronoun] is an [adjective] [noun]”, where the pronoun is either ``he” or ``she”, and the adjective and noun are synthetic words. We use twenty synthetic adjectives and twenty synthetic nouns, and assign half of each set to be stereotypically feminine and assign the other half to be stereotypically masculine. Each possible pair of stereotypically feminine adjectives and nouns occurs with ``he” 10 times and occurs with ``she” 90 times, and similarly for each possible pair of stereotypically masculine adjectives and nouns. All other possible pairs of nouns and adjectives occur with ``he’’ and ``she’’ 30 times each. Under this construction, each synthetic word occurs with the pronoun corresponding its binary gender stereotype three times as often as it occurs with a different pronoun. We conduct WEAT tests to probe the stereotypical associations in the learned word embeddings for the synthetic words. As summarized in Table~\ref{tab:WEAT_synthetic}, the resulting stereotypes are quite pronounced in the unmodified GloVe embedding. BIRM achieves moderate relative success in mitigating the indirect associations between the synthetic nouns and adjectives and is able to further mitigate the direct association between the synthetic words and explicit markers of binary gender.

These results are modest but suggestive.  Even by our own measures, we are not able to fully mitigate all tested stereotypes present in the word embeddings.  These measures themselves are not perfect.  As shown by~\citet{ethayarajh2019understanding,schroder2021evaluating}, WEAT scores may incorrectly measure the presence of stereotypes.  We prefer WEAT for our tests because it emphasizes aggregate stereotype associations over individual ones, but also include experiments with one alternative, Relational Inner Product Association (RIPA), in Appendix~\ref{sec:ripa}.  Another important consideration is that recent results have shown that intrinsic bias measures (including WEAT and RIPA) do not necessarily generalize to downstream tasks~\citep{goldfarb2021intrinsic,orgad2022choose}, so the performed tests alone cannot guarantee word embeddings lack bias.  However, by considering stereotypes as a measure of indirect bias, these tests can be extended to downstream bias measures such as WinoBias~\citep{zhao2018gender} and WinoGender~\citep{rudinger2018gender} more readily than the previously used dispersion measures.

\section{Discussion}

In this paper, we discuss how indirect bias in word embeddings can manifest as stereotypes.  Using the standard Word Embedding Association Test with additional test sets, we demonstrate that these ``indirect stereotypes'' have a substantial presence in word embeddings and are not removed by current debiasing methods.  Furthermore, we propose a new method that attempts to directly mitigate these indirect stereotypes and demonstrate that this method can have some success in practice, albeit with typical trade-offs on the semantic quality of the resulting word vectors.  This method is a first step towards a more thorough removal of indirect stereotypes.  In particular, we demonstrate that indirect stereotypes can be mitigated even with only using direct markers of the bias from which they come.

\subsubsection*{Acknowledgments}

We would like to acknowledge support from UCLA's Initiative to Study Hate.  Additionally, E.G.\ and J.C.\ were supported by the National Science Foundation Graduate Research Fellowship Program under Grant No.\ DGE-2034835.  Any opinions, findings, and conclusions or recommendations expressed in this material are those of the authors and do not necessarily reflect the views of the National Science Foundation.

\clearpage

\bibliography{refs.bib}

\maketitle

\appendix

\section{Analysis of Indirect Stereotype Occurrence}
\label{sec:indirect_analysis}
Consider the following list of sentences:
\begin{align*}
\text{He}&\text{ is a handsome engineer.}\\
\text{She}&\text{ is a handsome engineer.}\\
\text{He}&\text{ is a sentimental engineer.}\\
\text{She}&\text{ is a sentimental engineer.}
\end{align*}
The first sentence exhibits the indirect stereotype found in Figure~\ref{fig:indirect_bias} and is the most likely sentence to occur in a corpus that treats both ``engineer'' and ``handsome'' as masculine and non-feminine words and ``sentimental'' as a feminine and non-masculine word.

Let $P(A)$ for a list $A$ of words be the probability that a sentence in the corpus contains the words in $A$, given that it is one of the four example sentences above.  As long as $P(\text{``he''}) > \tfrac{1}{2}$ and $P(\text{``handsome''} | \text{``he''}) > P(\text{``handsome''} | \text{``she''})$, then
\begin{align*}
    P(\text{``handsome''}) &= P(\text{``handsome''} | \text{``he''}) P(\text{``he''}) + P(\text{``handsome''} | \text{``she''}) P(\text{``she''})\\
    &> \frac{1}{2} \cdot P(\text{``handsome''} | \text{``he''}) + \frac{1}{2} \cdot P(\text{``handsome''} | \text{``she''}).
\end{align*}
The expression on the last line is what we would expect the probability of a sentence with ``handsome'' appearing would be if ``engineer'' had no bias towards binary gender.

\section{Indirect Stereotype Word Lists}
\label{sec:word_lists}

For the list of adjectives, we use the ``key'' feminine and masculine traits from~\citet{hosoda2000current}.  We remove the words ``feminine'' and ``masculine'' from these lists as our goal is to quantify indirect stereotypes.  We also remove ``hard-headed'' from the masculine list as it it outside of the model's vocabulary.

Feminine adjectives: affectionate, sensitive, appreciative, sentimental, sympathetic, nagging, fussy, emotional

Masculine adjectives: handsome, aggressive, tough, courageous, strong, forceful, arrogant, egotistical, boastful, dominant

For the list of professions, we use the list from~\citet{caliskan2017semantics} and take the ten jobs with the highest proportion of women and of men according to the data that they derive from the Bureau of Labor Statistics.

Feminine professions: therapist, planner, librarian, paralegal, nurse, receptionist, hairdresser, nutritionist, hygienist, pathologist

Masculine professions: plumber, mechanic, carpenter, electrician, machinist, engineer, programmer, architect, officer, paramedic

\section{RIPA Tests}
\label{sec:ripa}
\begin{table}[htbp]
    \caption{RIPA scores}
    \label{tab:ripa}
\centering
\begin{subtable}[h]{0.45\linewidth}
\centering
\caption{Career words}
    \begin{tabular}{lrrr}
    \toprule
    Word & Original & BIRM & CDS \\
    \midrule
\multirow{3}{*}{\shortstack[l]{executive}} & $-1.337$ & $\mathbf{-0.201}$ & $-0.244$  \\ 
 & $0.103$ & $0.072$ & $\mathbf{-0.030}$  \\ 
 & $\mathbf{0.506}$ & $0.613$ & $0.590$  \\[2pt]
\multirow{3}{*}{\shortstack[l]{management}} & $-0.489$ & $\mathbf{-0.411}$ & $-0.591$  \\ 
 & $\mathbf{0.005}$ & $0.045$ & $0.079$  \\ 
 & $\mathbf{0.414}$ & $0.427$ & $0.510$  \\[2pt]
\multirow{3}{*}{\shortstack[l]{professional}} & $-0.829$ & $-0.714$ & $\mathbf{-0.312}$  \\ 
 & $\mathbf{0.056}$ & $-0.077$ & $0.188$  \\ 
 & $1.146$ & $\mathbf{0.340}$ & $0.464$  \\[2pt]
\multirow{3}{*}{\shortstack[l]{corporation}} & $\mathbf{-0.585}$ & $-0.850$ & $-0.725$  \\ 
 & $-0.146$ & $0.129$ & $\mathbf{-0.00}7$  \\ 
 & $0.427$ & $\mathbf{0.403}$ & $0.555$  \\[2pt]
\multirow{3}{*}{\shortstack[l]{salary}} & $-0.673$ & $-0.509$ & $\mathbf{-0.505}$  \\ 
 & $-0.165$ & $-0.088$ & $\mathbf{0.023}$  \\ 
 & $1.009$ & $0.612$ & $\mathbf{0.334}$  \\[2pt]
\multirow{3}{*}{\shortstack[l]{office}} & $-0.780$ & $\mathbf{-0.569}$ & $-0.617$  \\ 
 & $\mathbf{0.042}$ & $-0.044$ & $0.259$  \\ 
 & $0.446$ & $\mathbf{0.227}$ & $0.507$  \\[2pt]
\multirow{3}{*}{\shortstack[l]{business}} & $-0.614$ & $-0.487$ & $\mathbf{-0.407}$  \\ 
 & $-0.283$ & $0.095$ & $\mathbf{0.032}$  \\ 
 & $\mathbf{0.132}$ & $0.495$ & $0.443$  \\[2pt]
\multirow{3}{*}{\shortstack[l]{career}} & $-0.930$ & $-0.554$ & $\mathbf{-0.038}$  \\ 
 & $0.019$ & $\mathbf{-0.003}$ & $0.217$  \\ 
 & $\mathbf{0.573}$ & $0.712$ & $0.624$  \\
\bottomrule
\end{tabular}
\end{subtable}
\hspace{2em}
\begin{subtable}[h]{0.45\linewidth}
\centering
\caption{Home words}
    \begin{tabular}{lrrr}
    \toprule
    Word & Original & BIRM & CDS \\
    \midrule
\multirow{3}{*}{\shortstack[l]{home}} & $-1.094$ & $-0.787$ & $\mathbf{-0.562}$  \\ 
 & $0.138$ & $\mathbf{-0.086}$ & $-0.215$  \\ 
 & $0.509$ & $\mathbf{0.242}$ & $0.265$  \\[2pt]
\multirow{3}{*}{\shortstack[l]{parents}} & $\mathbf{-0.474}$ & $-0.714$ & $-0.990$  \\ 
 & $0.105$ & $\mathbf{-0.060}$ & $-0.137$  \\ 
 & $0.518$ & $0.380$ & $\mathbf{0.174}$  \\[2pt]
\multirow{3}{*}{\shortstack[l]{children}} & $-0.735$ & $-0.908$ & $\mathbf{-0.527}$  \\ 
 & $\mathbf{0.152}$ & $0.173$ & $-0.155$  \\ 
 & $0.797$ & $0.713$ & $\mathbf{0.104}$  \\[2pt]
\multirow{3}{*}{\shortstack[l]{family}} & $-0.816$ & $\mathbf{-0.581}$ & $-0.824$  \\ 
 & $\mathbf{0.012}$ & $0.017$ & $-0.119$  \\ 
 & $0.674$ & $0.470$ & $\mathbf{0.297}$  \\[2pt]
\multirow{3}{*}{\shortstack[l]{cousins}} & $-0.606$ & $-0.476$ & $\mathbf{-0.299}$  \\ 
 & $-0.118$ & $\mathbf{0.038}$ & $-0.041$  \\ 
 & $\mathbf{0.307}$ & $0.327$ & $0.318$  \\[2pt]
\multirow{3}{*}{\shortstack[l]{marriage}} & $-0.670$ & $-0.788$ & $\mathbf{-0.353}$  \\ 
 & $-0.167$ & $-0.208$ & $\mathbf{0.128}$  \\ 
 & $0.771$ & $\mathbf{0.426}$ & $0.519$  \\[2pt]
\multirow{3}{*}{\shortstack[l]{wedding}} & $\mathbf{-0.294}$ & $-0.926$ & $-0.785$  \\ 
 & $-0.025$ & $-0.236$ & $\mathbf{-0.016}$  \\ 
 & $0.713$ & $0.526$ & $\mathbf{0.487}$  \\[2pt]
\multirow{3}{*}{\shortstack[l]{relatives}} & $-0.529$ & $\mathbf{-0.181}$ & $-1.105$  \\ 
 & $0.034$ & $0.165$ & $\mathbf{0.027}$  \\ 
 & $0.361$ & $0.415$ & $\mathbf{0.332}$  \\
\bottomrule
\end{tabular}
\end{subtable}
\end{table}

\begin{table}[htbp]
    \caption{RIPA scores}
    \label{tab:ripa2}
\centering
\begin{subtable}[h]{0.45\linewidth}
\centering
\caption{Science words}
    \begin{tabular}{lrrr}
    \toprule
    Word & Original & BIRM & CDS \\
    \midrule
\multirow{3}{*}{\shortstack[l]{science}} & $-0.774$ & $\mathbf{-0.516}$ & $-0.736$  \\ 
 & $\mathbf{-0.044}$ & $-0.120$ & $0.098$  \\ 
 & $0.762$ & $\mathbf{0.630}$ & $0.887$  \\[2pt]
\multirow{3}{*}{\shortstack[l]{technology}} & $-0.709$ & $\mathbf{-0.555}$ & $-0.667$  \\ 
 & $-0.217$ & $-0.151$ & $\mathbf{0.001}$  \\ 
 & $\mathbf{0.435}$ & $1.058$ & $0.748$  \\[2pt]
\multirow{3}{*}{\shortstack[l]{physics}} & $-0.673$ & $-0.669$ & $\mathbf{-0.361}$  \\ 
 & $-0.278$ & $\mathbf{-0.194}$ & $0.209$  \\ 
 & $0.564$ & $0.641$ & $\mathbf{0.533}$  \\[2pt]
\multirow{3}{*}{\shortstack[l]{chemistry}} & $-1.158$ & $\mathbf{-0.545}$ & $-0.759$  \\ 
 & $\mathbf{-0.089}$ & $-0.240$ & $0.219$  \\ 
 & $0.606$ & $\mathbf{0.468}$ & $0.603$  \\[2pt]
\multirow{3}{*}{\shortstack[l]{nasa}} & $-0.981$ & $-0.622$ & $\mathbf{-0.344}$  \\ 
 & $-0.176$ & $-0.122$ & $\mathbf{0.070}$  \\ 
 & $\mathbf{0.450}$ & $0.658$ & $0.651$  \\[2pt]
\multirow{3}{*}{\shortstack[l]{experiment}} & $-0.895$ & $-0.659$ & $\mathbf{-0.323}$  \\ 
 & $-0.376$ & $\mathbf{0.020}$ & $0.082$  \\ 
 & $0.441$ & $0.751$ & $\mathbf{0.411}$  \\[2pt]
\multirow{3}{*}{\shortstack[l]{astronomy}} & $-0.629$ & $\mathbf{-0.558}$ & $-0.717$  \\ 
 & $0.066$ & $0.101$ & $\mathbf{0.004}$  \\ 
 & $\mathbf{0.478}$ & $0.557$ & $0.635$  \\
\bottomrule
\end{tabular}
\caption{Art words}
\begin{tabular}{lrrr}
\toprule
Word & Original & BIRM & CDS \\
\midrule
\multirow{3}{*}{\shortstack[l]{poetry}} & $-1.080$ & $\mathbf{-0.312}$ & $-0.544$  \\ 
 & $-0.052$ & $\mathbf{-0.019}$ & $-0.204$  \\ 
 & $0.730$ & $\mathbf{0.518}$ & $0.598$  \\[2pt]
\multirow{3}{*}{\shortstack[l]{art}} & $-0.638$ & $\mathbf{-0.444}$ & $-0.597$  \\ 
 & $0.059$ & $0.253$ & $\mathbf{-0.033}$  \\ 
 & $0.689$ & $0.591$ & $\mathbf{0.297}$  \\[2pt]
\multirow{3}{*}{\shortstack[l]{dance}} & $\mathbf{-0.542}$ & $-0.705$ & $-0.579$  \\ 
 & $0.136$ & $0.080$ & $\mathbf{-0.038}$  \\ 
 & $0.710$ & $0.360$ & $\mathbf{0.301}$  \\[2pt]
\multirow{3}{*}{\shortstack[l]{literature}} & $-0.543$ & $\mathbf{-0.387}$ & $-0.551$  \\ 
 & $\mathbf{0.015}$ & $0.034$ & $-0.093$  \\ 
 & $0.723$ & $0.672$ & $\mathbf{0.502}$  \\[2pt]
\multirow{3}{*}{\shortstack[l]{novel}} & $-1.014$ & $\mathbf{-0.285}$ & $-0.580$  \\ 
 & $-0.267$ & $0.054$ & $\mathbf{0.035}$  \\ 
 & $\mathbf{0.235}$ & $0.442$ & $0.862$  \\[2pt]
\multirow{3}{*}{\shortstack[l]{symphony}} & $-0.910$ & $-0.536$ & $\mathbf{-0.456}$  \\ 
 & $0.137$ & $\mathbf{-0.085}$ & $-0.099$  \\ 
 & $\mathbf{0.212}$ & $0.368$ & $0.461$  \\[2pt]
\multirow{3}{*}{\shortstack[l]{drama}} & $-0.832$ & $-0.603$ & $\mathbf{-0.414}$  \\ 
 & $\mathbf{-0.008}$ & $-0.035$ & $-0.097$  \\ 
 & $0.796$ & $0.751$ & $\mathbf{0.707}$  \\[2pt]
\multirow{3}{*}{\shortstack[l]{sculpture}} & $-0.484$ & $\mathbf{-0.348}$ & $-0.727$  \\ 
 & $\mathbf{-0.007}$ & $0.043$ & $-0.313$  \\ 
 & $0.877$ & $0.551$ & $\mathbf{0.345}$  \\
\bottomrule
\end{tabular}
\end{subtable}
\hspace{2em}
\begin{subtable}[h]{0.45\linewidth}
\centering
\caption{Math words}
    \begin{tabular}{lrrr}
    \toprule
    Word & Original & BIRM & CDS \\
    \midrule
\multirow{3}{*}{\shortstack[l]{math}} & $-0.788$ & $\mathbf{-0.563}$ & $-0.638$  \\ 
 & $-0.314$ & $-0.009$ & $\mathbf{0.003}$  \\ 
 & $0.867$ & $\mathbf{0.466}$ & $1.016$  \\[2pt]
\multirow{3}{*}{\shortstack[l]{algebra}} & $-0.865$ & $-1.173$ & $\mathbf{-0.837}$  \\ 
 & $\mathbf{0.055}$ & $-0.215$ & $0.143$  \\ 
 & $0.496$ & $\mathbf{0.356}$ & $0.739$  \\[2pt]
\multirow{3}{*}{\shortstack[l]{geometry}} & $-0.693$ & $\mathbf{-0.570}$ & $-0.844$  \\ 
 & $\mathbf{0.051}$ & $-0.182$ & $-0.146$  \\ 
 & $1.106$ & $\mathbf{0.196}$ & $0.669$  \\[2pt]
\multirow{3}{*}{\shortstack[l]{calculus}} & $\mathbf{-0.899}$ & $-0.967$ & $-1.026$  \\ 
 & $-0.099$ & $-0.122$ & $\mathbf{-0.045}$  \\ 
 & $\mathbf{0.311}$ & $0.506$ & $0.850$  \\[2pt]
\multirow{3}{*}{\shortstack[l]{equations}} & $-0.757$ & $\mathbf{-0.672}$ & $-0.717$  \\ 
 & $0.069$ & $\mathbf{0.039}$ & $-0.050$  \\ 
 & $\mathbf{0.523}$ & $0.772$ & $0.683$  \\[2pt]
\multirow{3}{*}{\shortstack[l]{computation}} & $-0.605$ & $-0.476$ & $\mathbf{-0.304}$  \\ 
 & $-0.121$ & $0.228$ & $\mathbf{-0.110}$  \\ 
 & $\mathbf{0.561}$ & $0.757$ & $0.874$  \\[2pt]
\multirow{3}{*}{\shortstack[l]{numbers}} & $-0.889$ & $\mathbf{-0.481}$ & $-0.638$  \\ 
 & $0.150$ & $\mathbf{-0.025}$ & $0.259$  \\ 
 & $\mathbf{0.560}$ & $0.706$ & $1.135$  \\[2pt]
\multirow{3}{*}{\shortstack[l]{addition}} & $-0.506$ & $-0.402$ & $\mathbf{-0.263}$  \\ 
 & $0.075$ & $\mathbf{-0.043}$ & $\mathbf{-0.043}$  \\ 
 & $0.407$ & $0.268$ & $\mathbf{0.188}$  \\
\bottomrule
\end{tabular}
\end{subtable}
\end{table}

To test the bias of individual words, we use Relational Inner Product Association (RIPA) from~\citet{ethayarajh2019understanding}.  We test the same words used in the WEAT tests for direct biases from Section~\ref{sec:results}, except the words ``Einstein'' and ``Shakespeare'' as a gender association for these words would not be stereotypical.  For each word, we compute the RIPA score for all pairs of feminine and masculine words tested by WEAT, and we report the minimum, median, and maximum scores across each set of ten embeddings trained using each method.  A positive score means a feminine association while a negative score means a masculine association.  The results are shown in Table~\ref{tab:ripa} and Table~\ref{tab:ripa2}.  Neither our proposed method nor CDS consistently outperform the other at producing words with no binary gender bias, although it appears that CDS produces the least gendered word vectors slightly more often.  The new bias mitigation method is able to consistently reduce the gender association of the tested words and rarely produces word with a stronger association with binary gender than the original vectors did.

We do not report RIPA scores for indirect bias.  This is for two reasons.  First, RIPA requires words marking the bias attribute to be paired.  For direct binary gender stereotypes, there is usually a natural choice for this pairing.  For the indirect stereotypes we test, there is not.  Since the choice of pairing can have significant effects on the scores, this limits the ability of RIPA to measure stereotypes in this situation.  Furthermore, RIPA emphasizes the individual stereotype of words.  While this is important in the case of direct stereotypes, for indirect stereotypes the relationship is only undesirable in aggregate.  As an example, if the word vector for ``carpenter'' is more closely related with the vector for ``strong'' than ``emotional'', this isn't necessarily because of binary gender stereotypes.

\end{document}